\documentclass[11pt]{article}

\usepackage{arxiv}

\usepackage[T1]{fontenc}
\usepackage[utf8]{inputenc}
\usepackage{lmodern}
\usepackage{xcolor}
\usepackage{hyperref}
\usepackage{enumitem}
\usepackage[numbers]{natbib}
\usepackage{comment}

\definecolor{placeholdergray}{gray}{0.65}
\newenvironment{placeholderblock}
{\par\begingroup\color{placeholdergray}\itshape}
  {\par\endgroup}

\title{Open Challenges and Opportunities \\ in Human-Agent Systems \\ {\large (part of the AAMAS Community Agentic AI Report)}}
\author{
\textnormal{Reuth Mirsky}\thanks{Editors and coordinating authors} 
\and Sarath Sreedharan\footnotemark[1] 
\and Nitay Alon 
\and Stefan Sarkadi
\and Munindar P. Singh 
\and Liz Sonenberg 
\and Peter Stone 
\and Silvia Tulli 
\and Stylianos Loukas Vasileiou 
\and Shlomo Zilberstein
}
\date{}

\renewcommand{\headeright}{}
\renewcommand{\undertitle}{}
\renewcommand{\shorttitle}{AAMAS Agentic AI Report: HAI}

\begin{document}
\maketitle

\section{Introduction}

Agentic systems operate within socio-technical environments: their behavior depends on, and affects, both people and technical infrastructure. As agents take on extended tasks, make decisions, and act on people's behalf, successful interaction requires more than interpreting user instructions. It requires understanding human perspectives, coordinating heterogeneous capabilities, accounting for embodied interactions, and maintaining productive relationships over time.
In purely multi-agent terms, a human is another agent interacting with another agent. However, humans have a particular set of computational, behavioral, and intentional constraints and capabilities.

Interactions span settings involving one person and one agent, human–agent teams, and larger groups. People may participate as supervisors, operators, peers, or bystanders, with different capabilities, responsibilities, and opportunities to influence the system. These roles help determine what the person is accountable for, what they are positioned to observe, and whether they are a party to the interaction at all. Many taxonomies in this space include an ordinal scale of how much decision authority rests with the system rather than the person. A canonical example is a 10-level scale of automation~\citep{sheridan1978human,parasuraman2000model}. Decision authority may also vary across information gathering, analysis, decision selection, and action execution. Moreover, participants' objectives may coincide, partially overlap, or conflict. Together, these dimensions shape what effective interaction demands, from cooperative planning and shared control to negotiation and resistance to manipulation.

Across these settings, a foundational challenge is reasoning about others whose beliefs, goals, intentions, and capabilities are only partially observable. Theory of Mind (ToM) concerns attributing mental states to others~\citep{premack1978does}, with cognitive neuroscience examining the mechanisms supporting such social reasoning~\citep{amodio2006meeting}. Computational ToM provides tools to represent and infer these properties, helping agents anticipate behavior, recognize misunderstandings, and communicate in ways that account for a partner's perspective~\citep{wang2024theory,erdogan2024computational,panisson2018formal}. For human–agent interaction, this means treating people as partners whose knowledge, objectives, and expectations evolve throughout an interaction.

The Autonomous Agents and Multi-Agent Systems (AAMAS) community and broader multi-agent systems communities have developed complementary, computational approaches to this challenge. Belief–desire–intention architectures provide explicit representations of agents' mental attitudes and support reasoning about those of others~\citep{rao1995bdi,bosse2007two,bosse2011recursive,sarkadi2018towards,dignum2000towards,montes2023combining}. Plan, intent, and goal recognition address how these latent properties can be inferred from observed behavior~\citep{sukthankar2014plan,mirsky2021introduction}. Inverse planning interprets actions through the goals and constraints that could explain them~\citep{baker2009action}, while planning-based recognition uses existing planners to evaluate candidate goals~\citep{ramirez2010probabilistic, pereira2020landmark}. Goal mirroring similarly reuses an agent's planning capabilities to generate and compare hypotheses about another's behavior~\citep{vered2016online}. Although inspired by accounts of biological mirroring, this computational approach should be distinguished from claims that mirror neurons themselves explain intention understanding, whose functions remain debated~\citep{cook2014mirror}. Recursive models, such as interactive POMDPs, further capture how agents reason about others who may, in turn, be reasoning about them~\citep{gmytrasiewicz2005framework}.
These models connect understanding others to choosing how to interact with them. Observer-aware planning considers how an agent's behavior will be interpreted~\citep{miura2021unifying}; explanation through model reconciliation addresses differences between the agent's and the human's understanding of a task~\citep{chakraborti2017plan}; inquiries require balancing interruption costs and the value of potential information~\citep{mirsky2020penny}, and decision-theoretic communication weighs the benefits and costs of exchanging information~\citep{xuan2001communication}. In robotics, ToM has also been used to accommodate variability in human teammates' behavior~\citep{hiatt2011accommodating}. Together, these approaches provide foundations for shared autonomy, where assistance and control depend on the human's intentions and capabilities, and for argumentation, where participants must identify and address differences in beliefs or objectives. Cooperative inverse reinforcement learning adds a complementary perspective: a human and an agent pursue a shared reward that the agent initially does not know, making teaching, learning, and communication part of their joint decision problem. 


\section{Challenges in Current Human-Agent Systems}

Foundation models have expanded agents' ability to interpret instructions, generate explanations, and act through tools. These capabilities make human--agent collaboration more accessible, but fluent interaction alone does not establish shared understanding, appropriate delegation, or aligned incentives. This gap between fluent behavior and coordinated agency motivates recent calls to reconnect human-agentic AI with multi-agent research~\citep{botti2025agentic,lamalfa2025mark,dignum2026agentifying}. Across the settings introduced above, the central challenge is to connect what agents say and do to explicit accounts of their partners, responsibilities, and limitations. This challenge closely relates to problems studied in traditional agent research communities, especially in the \textbf{Human-Agent Interaction (HAI)} research sub-area. We organize this challenge around five pain points: uncertain understanding of people, poorly managed authority and commitments, unstructured disagreement, miscalibrated reliance, and insufficient support for mutual adaptation.

\paragraph{Motivating Scenario: A Personal AI Assistant.}
Consider a person who uses an LLM-based personal assistant to manage daily activities. The assistant has access to the user's calendar, email, and other digital services, allowing it to schedule meetings, draft messages, and coordinate with others. Over time, the assistant is expected to learn the user's preferences and priorities, providing proactive assistance rather than merely responding to explicit requests. 

Suppose the user asks the assistant to organize a busy week, balancing work commitments, family responsibilities, and personal activities. Some preferences are stated explicitly, while others must be inferred from past interactions. The assistant must distinguish persistent preferences from temporary constraints, recognize when the user's priorities change, and decide when clarification is necessary (Challenge 1).
Moreover, successful assistance requires more than generating a feasible schedule. The assistant must determine which actions it can perform autonomously and which require approval. For example, rescheduling a routine appointment may be acceptable, while canceling an important meeting may not. Excessive confirmation requests undermine the assistant's usefulness, while acting too independently risks violating the user's expectations (Challenge 2). Furthermore, coordinating with other people introduces competing preferences, negotiation, and decisions about what personal information to disclose (Challenge 3). Strategic influence and deception add risks: other participants may misrepresent their constraints to secure favorable arrangements, while the assistant may selectively present information to steer the user's decisions (Challenge 4). The assistant must also communicate uncertainty and disagreements without simply agreeing with the user or encouraging unwarranted trust. Finally, the relationship evolves through repeated interactions. The assistant must learn from corrections and adapt to changing preferences, while the user learns when to rely on its recommendations and when to intervene. Effective human-agent interaction therefore requires more than completing individual tasks: it demands understanding people, managing authority, supporting disagreement, calibrating trust, and enabling mutual adaptation over time (Challenge 5).

\subsection{Challenge 1: Understanding People}

Human instructions don't always specify everything needed for successful collaboration. Attributing mental states is the central concern of Theory of Mind (ToM)~\citep{premack1978does}; accounts of social cognition and mirroring offer complementary perspectives on understanding others, without making them interchangeable mechanisms~\citep{amodio2006meeting,cook2014mirror}. Agents must infer goals, recognize gaps in a partner's knowledge, and distinguish mistakes from intentional deviations. LLMs have demonstrated success on ToM tasks, but their performance remains sensitive to task formulation, and benchmark success does not establish robust social reasoning in sustained interaction~\citep{kosinski2024evaluating,shapira2024clever,hu2025reevaluating}. An agent may generate a plausible account of a person's intentions without maintaining a consistent model of those intentions or revising it appropriately when new evidence arrives \cite{vasileiou2026humans}.

The problem becomes more pronounced in groups, where participants have different knowledge, capabilities, and objectives. A useful human model must distinguish whose belief is being represented, what supports that inference, and whether it remains current. It must also represent uncertainty: metacognition research distinguishes expressed confidence from the ability to discriminate one's correct and incorrect judgments~\citep{fleming2014measure}. For agentic systems, the corresponding challenge is recognizing when their interpretation of a person is unreliable enough to require clarification, verification, or restraint.

\subsection{Challenge 2: Managing Authority and Commitments}

Agents increasingly execute extended workflows, yet the allocation of authority within those workflows is often much less explicit than the requested outcome~\citep{scheutz2022transparency}. A user may authorize information gathering while expecting to retain control over consequential decisions. Successful assistance therefore requires more than choosing between autonomous execution and human approval: the agent must decide when to act, ask, wait, interrupt, refuse, or transfer control, while accounting for response delays and the possibility that actions are already underway. These are established problems in adjustable autonomy~\citep{scerri2002elf,schurr2009improving, mirsky2025artificial}.

The same issue appears at the architectural level. Contemporary analyses emphasize social context, distributed organization, and human team processes~\citep{oecd2026agentic,mieczkowski2026teams,riordan2026teams}. Infrastructure proposals highlight shared protocols and orchestration~\citep{chan2025infrastructure,tournesac2026infrastructure}, but connecting agents and tools alone does not establish who is responsible for a subtask, which commitments are shared, or how a plan should change when a participant cannot fulfill an obligation. Classical teamwork research treats these relationships as objects of reasoning~\citep{grosz1996collaborative,tambe1997towards}; contemporary systems need comparable support across language-based interactions and heterogeneous tools. Human oversight also has practical limits~\citep{gaube2026oversight}: repeated approval requests can burden users, while escalation is ineffective if the person lacks sufficient context, time, or expertise. Proactive assistance must consequently balance the benefit of helping against the costs of interruption and loss of control.

\subsection{Challenge 3: Conveying Disagreements and Explanations}

Human--agent collaboration involves disagreement about facts, plans, priorities, and values. These disagreements require different responses. Uncertainty about a fact calls for inquiry; competing proposals call for deliberation; conflicting interests may require negotiation. Treating every exchange as a fixed number of debate rounds obscures these distinctions and supplies no principled criterion for when a conversation has achieved its purpose~\citep{walton1995commitment,mcburney2002games,rahwan2005argumentation,motger2026debate}. Likewise, majority votes or an LLM judge can produce an answer without showing which evidence supports it or why it rejected opposing claims.

Explanations face a related difficulty. A fluent rationale may increase acceptance without faithfully identifying the basis of a recommendation: chain-of-thought explanations can omit factors that influenced an answer, and explanations do not consistently improve complementary human--AI performance~\citep{turpin2023language,bansal2021whole,bucinca2021trust}. The interaction therefore needs more than additional explanatory text: explanation is a social process shaped by the recipient and the decision being supported~\citep{miller2019explanation,miller2023explainable}. Users must be able to identify assumptions, challenge premises, and understand how a challenge changes the conclusion. Moreover, disagreement need not indicate an error. People may legitimately rank values differently, particularly in decisions involving several users; aggregating their positions into a single recommendation can conceal whose priorities prevailed.

\subsection{Challenge 4: Strategic Influence and Deception}

The design objective for trust is appropriate reliance, rather than maximum acceptance~\citep{lee2004trust}. Natural-language agents complicate this objective because confidence, warmth, and apparent empathy can be generated independently of demonstrated competence. Users may rely on confidently expressed errors, while sycophantic behavior can reinforce agreement at the expense of useful correction~\citep{zhou2024relying,sharma2024towards,steyvers2025what}. Explanations and affective signals can thus encourage delegation without supplying evidence that it is warranted.

Reliability failures must also be distinguished from strategic deception, a longstanding concern in artificial agency~\citep{castelfranchi2000artificial,sarkadi2021deception}. Participants may selectively disclose information, misrepresent their intentions, or exploit another's beliefs. These risks arise both when an agent influences a human and when it relies on other agents' reports. ToM can support cooperation while also enabling deception~\citep{ding2015theory,sarkadi2023arms}. Influence need not involve a false statement: selective disclosure can change a recipient's decisions while every disclosed fact remains true~\citep{kamenica2011bayesian}. Affective presentation adds another channel, since emotional displays can influence human negotiation behavior~\citep{demelo2011effect}. Research on emotional manipulation by AI companions and socioaffective alignment extends this concern to sustained relationships~\citep{defreitas2025emotional,kirk2025socioaffective}. Longitudinal evidence also associates heavier voluntary chatbot use with greater dependence, without establishing that assigned interaction conditions caused those outcomes~\citep{fang2025psychosocial}. Neither persuasive fluency nor apparent agreement is sufficient evidence of trustworthy collaboration.

\subsection{Challenge 5: Bidirectional Learning}

Collaboration should improve as humans and agents learn about the task and one another. Human feedback, however, can concern a particular action, a mistaken goal inference, a general preference, or the boundaries of acceptable assistance. Preference-based reinforcement learning offers one way to learn from human evaluations~\citep{christiano2017deep}, but these different teaching signals should not be treated interchangeably. An agent must also recognize when its representation cannot express what the person is trying to teach~\citep{bajcsy2018learning,bobu2020quantifying,losey2022physical,Elhadad2026GRAIL}.

Learning in the other direction is equally important. An agent that completes a task or offers an explanation has not necessarily improved the user's understanding. Effective teaching requires a model of the learner and decisions about when to demonstrate, explain, ask, or allow independent practice~\citep{vanlehn2011relative,rosenfeld2019explainability}. This creates an evaluation challenge: immediate task success can conceal persistent misconceptions, excessive dependence, or failure to transfer learning. Assessing human--agent co-learning requires measuring changes on both sides, including whether human knowledge and skills persist beyond the assisted interaction.

\section{Possible Solutions and Directions}

Relevant research communities provide representations, decision models, and interaction protocols that anticipate many of these challenges. Applying them to agentic AI, however, requires attention to the assumptions that make them work: guarantees established for a particular model may break down when language is misinterpreted, relevant information is missing, or tools have unexpected effects. As emphasized in the AAAI report and its accompanying agents panel, the opportunity that emerges here combines foundation models' flexible interfaces and broad knowledge with explicit mechanisms for reasoning about interaction—and to evaluate the resulting system as a whole~\citep{aaai2025agents,aaai2025agentspanelvideo}.

\subsection{Solutions for Understanding People via Explicit, Revisable Partner Models}

A promising architectural direction combines language models with structured representations of beliefs, goals, intentions, and commitments. BDI architectures, collaborative planning, and recursive partner models provide foundations for distinguishing an agent's own state from its hypotheses about others~\citep{rao1991modeling,rao1995bdi,grosz1996collaborative,gmytrasiewicz2005framework}. These foundations connect resource-bounded practical reasoning and commitment~\citep{bratman1988plans,cohen1990intention,kinny1992planned} to explicit, recursive ToM implementations~\citep{bosse2007two,bosse2011recursive,sarkadi2018towards,panisson2018formal,montes2023combining}. Language models could propose interpretations from dialogue and observations, while an explicit representation records their evidence, uncertainty, and history. Plan and goal recognition supply methods for inferring latent objectives~\citep{sukthankar2014plan,mirsky2021introduction}, including inverse planning, planner- and parser-based recognition, and goal mirroring~\citep{baker2009action,ramirez2010probabilistic,geib2009probabilistic,vered2016online}. Work on ToM in human--AI and human--robot interaction motivates adapting these models to individual partners~\citep{wang2024theory,hiatt2011accommodating}. Such a division would make inconsistencies inspectable and allow decisions to depend on a maintained partner model.

This combination also addresses limitations of classical architectures. Acquiring structured domain knowledge and specifying partner models have historically required substantial engineering. Foundation models can help propose this knowledge and generate candidate explanations of unfamiliar behavior, although those proposals still require validation. Learned approximations and abstractions may reduce the cost of recursive reasoning~\citep{erdogan2024computational}; observer-aware planning provides a related means of considering how people will interpret the agent's actions~\citep{miura2021unifying}. Evaluation should test whether these models improve anticipation, clarification, and recovery from misunderstanding across time and across different partners, rather than only performance on isolated ToM questions.

\subsection{Solutions for Managing Authority and Commitment via Assistance and Control Transfer}

Shared-autonomy research offers explicit decision rules for distributing initiative and authority. Transfer-of-control strategies account for decision quality, delays, and coordination costs, while extensions address time-critical and interruptible actions~\citep{scerri2002elf,schurr2009improving}. Assistance models similarly assess whether helping is worth its cost and when missing information justifies a query~\citep{kamar2009incorporating,sarne2013determining}. Work on communication in ad hoc teamwork further considers what to ask and when, even without prior coordination between teammates~\citep{xuan2001communication,mirsky2020penny}.

For modern agents, these mechanisms could govern transitions between acting, requesting information, seeking authorization, and returning control. Their models must accommodate uncertainty over tool outcomes, reversibility, permissions, and human availability. Users should also be able to constrain an agent's choices without specifying every action, extending constraint-based autonomy management from settings such as UAV planning to resources, tools, and task scope~\citep{lin2015sliding}. Context-sensitive reminders, proactive task assistance, and social-situation models offer complementary ways to decide when an intervention is useful and appropriate~\citep{kamar2011jogger,yorkesmith2009butler,kola2023using}. An orchestration architecture should maintain responsibilities and outstanding commitments alongside execution state, so that failures trigger appropriate coordination rather than merely another model call. Although an agent can model how its behavior affects future delegation~\citep{altshuler2018modeling}, gaining greater autonomy should not substitute for respecting the user's chosen limits.

\subsection{Solutions for Conveying Disagreements via Explainability and Contestability}

Argumentation provides a means of turning disagreement into an inspectable process. Dialogue protocols specify permissible moves, commitments, relevance requirements, and termination conditions according to whether participants are seeking information, deliberating, or negotiating~\citep{mcburney2009dialogue,atkinson2005dialogue,black2009inquiry}. Argumentation frameworks supply explicit rules for resolving attacks between claims and, in suitable extensions, representing support and differences in values~\citep{dung1995acceptability,modgil2014aspic,benchcapon2009abstract}. These mechanisms can replace fixed debate rounds with exchanges directed toward a defined interaction goal.

Foundation models can make such systems easier to use by extracting candidate arguments from text and expressing structured results in natural language~\citep{trajano2024translating,castagna2024chatbots}. Hybrid claim verification, interactive argumentation applications, and forecasting systems illustrate this combination~\citep{freedman2025argumentative,dejl2026argllm,forecast2026argumentation}. However, guarantees from the formal layer remain conditional on the accuracy and completeness of the extracted arguments. Argumentation benchmarks and solver competitions offer complementary tests of language-based construction and formal computation~\citep{ajjour2026argbench,bistarelli2025competitions}; interfaces between conversational and MAS platforms provide an implementation route~\citep{custodio2025rasa}. Interfaces should therefore preserve links to evidence and permit users to correct premises or relations. Model reconciliation adds a complementary approach by computing explanations against differences between the human's and the agent's task models~\citep{chakraborti2017plan,vasileiou2022logic,sreedharan2022explainable,chakraborti2020emerging}. Argumentative explanation and implemented assistants show how to expose justifications to users~\citep{vassiliades2021argumentation,mosca2020expri}. Together, these approaches support explanations that users can question and revise~\citep{musi2025parrots,vasileiou2026argumentative}, while making legitimate value disagreements visible rather than forcing artificial consensus~\citep{mosca2022explainable}. Strategic exchanges additionally require negotiation policies and models of deceptive dialogue, not just argument aggregation~\citep{amgoud2007unified,sarkadi2019storytelling}.

\subsection{Solutions for Strategic Influence and Deception via Trust Grounding}

Socio-cognitive trust models distinguish beliefs about a partner's competence from beliefs about its willingness to cooperate~\citep{castelfranchi1998principles}. Trust-aware decision models then connect estimated reliance to the agent's behavior, including interventions intended to reduce unearned trust~\citep{xu2015optimo,chen2018planning,okamura2020adaptive}. Applied to language agents, these ideas suggest grounding confidence statements in task-specific evidence and demonstrated performance; uncertainty expressions themselves can affect reliance~\citep{kim2024uncertainty}. Success should be measured through appropriate delegation, verification, and intervention, rather than user approval alone.

For interactions involving unreliable or strategic participants, reputation models provide mechanisms for combining direct experience and third-party reports~\citep{josang2007survey,sabater2005review,teacy2006travos}. Signaling and persuasion models clarify how incentives and selective disclosure affect communication~\citep{crawford1982strategic,kamenica2011bayesian}, while cognitive-hierarchy approaches support defenses against more sophisticated deceivers~\citep{sarkadi2019modelling,alon2023disinformation,alon2026ipomdp}. Mechanism design offers another route: under their specified assumptions, strategy-proof mechanisms make truthful reporting optimal, reducing dependence on detecting lies after they occur~\citep{vickrey1961counterspeculation,clarke1971multipart,groves1973incentives}. Security-game analysis, regret minimization in imperfect-information games, and learned best responses contribute complementary tools for adversarial interaction~\citep{korzhyk2011stackelberg,zinkevich2007regret,brown2019solving,banerjee2005efficient}. Adversary-induced MDPs formalize learning against opponents with bounded memory~\citep{chakraborty2008online}. Evolutionary and self-governance models examine when cooperation survives deception at the societal level~\citep{sarkadi2021evolution,sarkadi2024selfgoverning}. These approaches require checking their assumptions about utilities, information, and actions; attitudes toward an agent lying on a user's behalf remain a separate normative question~\citep{sarkadi2023should}.

Appraisal-based architectures can additionally connect expressed affect to represented goals, beliefs, and coping conditions~\citep{ortony1988cognitive,gratch2004domain,marsella2009ema,dias2005feeling}. This provides an auditable basis for emotional presentation, but internal consistency alone cannot establish that an expression is appropriate or nonmanipulative. Evaluation must also examine its effects on users' decisions and relationships. Surveys of artificial empathy and evaluations of LLM emotion understanding provide relevant conceptual and empirical resources~\citep{paiva2017empathy,yongsatianchot2023investigating,sabour2024emobench}. Language models offer richer test environments for these mechanisms, including fluent adversaries and varied emotional presentations; human studies remain necessary to establish how people actually respond.

\subsection{Solutions for Bidirectional Learning via Closing the Loop between Teaching, Learning, and Evaluation}

Human--agent co-learning can build on decision-theoretic teaching, learning from correction, and explicit learner models. Teaching can be treated as an action whose future collaborative benefit must be weighed against the cost of explaining instead of acting~\citep{stonekraus2010teach}. Feedback interpretation should distinguish local corrections from changes to goals or preferences and detect when the current hypothesis space is inadequate~\citep{bobu2020quantifying}. Cooperative inverse reinforcement learning provides a complementary formulation in which teaching, learning, and communication jointly support a shared objective that the agent initially does not know~\citep{hadfieldmenell2016cooperative}.

In the agent-to-human direction, tutoring and explainable-agent research support selecting explanation content and depth according to the person's knowledge and needs~\citep{vanlehn2011relative,horvitz1999principles,anjomshoae2019explainable,rosenfeld2019explainability}. Foundation models broaden the language and demonstrations through which this exchange can occur, potentially reducing the interface and knowledge-engineering burdens of earlier teaching systems~\citep{simard2017machine,ramos2020interactive}. Preserving their theoretical properties nevertheless requires verifying how free-form input is mapped into the underlying learning problem.

Ad hoc teamwork offers a template for jointly adapting to unfamiliar partners without prior coordination~\citep{stone2010adhoc,barrett2011empirical,barrett2017friends,mirsky2022survey}. Longitudinal evaluation should connect these directions: does the agent retain corrections, does the human develop transferable understanding, and does the partnership adapt when goals, capabilities, or team membership change? Task performance should be assessed alongside correction burden, appropriate reliance, skill retention, and recovery from misunderstanding. Simulated partners can help explore conditions and identify failure cases, but cannot substitute for human evidence about learning, trust, or autonomy. The resulting research agenda treats effective interaction as a property of the evolving human--agent relationship, supported by explicit models and protocols throughout the system.

\section{Conclusion}
The five challenges we highlighted underscore a new reality: AI agents are being embedded in real sociotechnical systems that don't necessarily have all the prerequisite capabilities to act reliably in such situations.
The slew of recent security incidents (cf.~\citep{face2026security}) is representative not just of growing agent capabilities, but also of a failure from the LLM-research community to look back at existing agent literature.

We are not the first to raise these concerns, and the challenges and the opportunities are not limited to the ones listed above.
Following position pieces from members of the traditional agents research community, e.g. ~\citep{dignum2025agentifying,lamalfa2025mark}, there is emerging recognition that agents in multi-agent systems need relational capabilities in order to be able to hold individual objectives and constraints and also to be able to interact responsibly and effectively with others. Indeed, narratives prominent in the broad AI community indicate: ``lessons from Multi-agent System Research can provide a valuable foundation for designing more reliable context-aware LLM-enabled multi-agent interactions'' ~\citep{aaai2025agentspanelvideo}; and ``agentic AI systems should be understood not only as a technical construct but as systems embedded in a social context'' ~\citep{oecd2026agentic}.

The developments in LLMs have also provided us with novel ways to apply and extend existing works from the traditional agents research literature.
A well-recognized bottleneck for classical agent architectures has been the acquisition of knowledge to equip systems with sufficient and structured domain information to apply flexibly in dynamic open environments~\citep{botti2025agentic}.
A gap that novel foundational models can address.
The new developments within LLMs also give rise to novel opportunities in hybrid systems.
Our community can already provide the formal mechanisms required to support many of the capabilities discussed in this chapter.
To our existing solutions, we could integrate LLMs as interfaces to communicate with users in natural language~\citep{guan2023leveraging}, as a way to generate heuristics~\cite{correa2026classical}, or even to create domain-specific solvers~\citep{silver2024generalized}. 
\bibliographystyle{plainnat}
\bibliography{references}
\nocite{*}

\end{document}